\documentclass[letterpaper, 10 pt, conference]{ieeeconf}  

\IEEEoverridecommandlockouts                              

\usepackage{amsmath,amssymb,amsfonts}
\usepackage{algorithmic}
\usepackage{subcaption}
\usepackage{graphicx,booktabs,multirow}
\usepackage{textcomp}
\usepackage{xcolor,soul}
\usepackage{color}

\usepackage{url}
\usepackage{fancyvrb,fancyhdr}
\usepackage{commath}
\usepackage{comment}
\usepackage{fleqn,multirow,wrapfig,lineno,hyperref}
\hypersetup{
    colorlinks=true,
    linkcolor=blue,
    filecolor=blue,      
    urlcolor=blue,
    bookmarksopen=true,
    citebordercolor=blue,
    filebordercolor=blue,
    linkbordercolor=blue,
    menubordercolor=blue,
    urlbordercolor=blue
}
\sethlcolor{yellow}

\definecolor{bronze}{rgb}{0.8, 0.5, 0.2}
\definecolor{blue}{rgb}{0, 0, 1}
\definecolor{green}{rgb}{0, 1, 0}
\definecolor{black}{rgb}{0, 0, 0}
\definecolor{red}{rgb}{1, 0, 0}

\newcommand{\myVspace}{\vspace{5 pt}}

\graphicspath{{figs/}}

\title{\LARGE \bf
Leveraging Inter-object Affordances for Efficient Planning\\in Contact-rich Tasks
}

\author{Pouya P. Niaz$^{1}$, Justus Piater$^{1}$, and Alejandro Agostini$^{1}$
\thanks{*This work was supported by the Austrian Science Fund (FWF) Project P36965 [DOI: 10.55776/P36965].}
\thanks{$^{1}$ P. P. N., J. P. and A. A. are with the Intelligent and Interactive Systems (IIS) research group, Department of Computer Science, University of Innsbruck, Technikerstraße 21a, 6020 Innsbruck, Austria.
        {\tt\small \{pouya.pourakbarian-niaz, justus.piater, alejandro.agostini\}@uibk.ac.at}}%
}

\begin{document}

\maketitle
\thispagestyle{empty}
\pagestyle{empty}

\begin{abstract}

Traditional task-and-motion planning (TAMP) approaches primarily focus on defining sequences of actions along with the necessary geometric and kinematic constraints to execute long-horizon tasks. 
However, their applicability in real-world settings is limited, as they typically assume simplified object models that overlook key physical properties critical for the successful execution of contact-rich tasks.
Moreover, they often use sub-symbolic reasoning during motion planning, which drastically increases planning time and decreases overall success rates.
We propose a method that leverages a TAMP approach, defining object-centric abstractions of execution constraints, called \emph{Unified} TAMP (U-TAMP), to execute robotic tasks involving interactions among objects with heterogeneous shapes, sizes, and materials.
Using a Vision-Language Model (VLM), we generate abstractions of inter-object affordances for characterizing physical interaction constraints between objects in contact-rich tasks, such as grasp and support constraints.
These constraints are used to enrich the U-TAMP planning domain to deal with objects with variable physical properties.
We perform experiments in simulated kitchen table organization scenarios and compare our results with those of the original U-TAMP, as well as a state-of-the-art VLM-based planner that leverages common sense knowledge of objects' affordances for plan generation.
Our approach achieves significantly higher planning success rates and improves planning times by one to two orders of magnitude compared to other methods.
\end{abstract}






\section{Introduction}
    \label{sec:intro}
%
%
Task and Motion Planning (TAMP) has enjoyed widespread growth for long-horizon planning in robotic manipulation tasks thanks to readily available AI planning tools and better learning-based methods~\cite{Garrett2021, Mansouri2021, Guo2023}.
Despite recent promising results, TAMP approaches still show limited performance in realistic settings, as execution constraints are typically defined using rudimentary object models. For instance, objects are often represented as lightweight, manageable cuboids or other simple shapes, and many relevant physical properties are ignored~\cite{Bidot2017, Orthey2024, Vu2024,garrett2018ffrob, garrett2020pddlstream, agostini2023unified}. 
This limits the ability to define all the necessary physical constraints for object manipulation and object-object interactions, and hinders the full exploitation of objects' affordances.

\begin{figure}[t!]
    \myVspace
	\centering
	\begin{subfigure}[t!]{0.635\columnwidth}
		\includegraphics[width=\columnwidth]{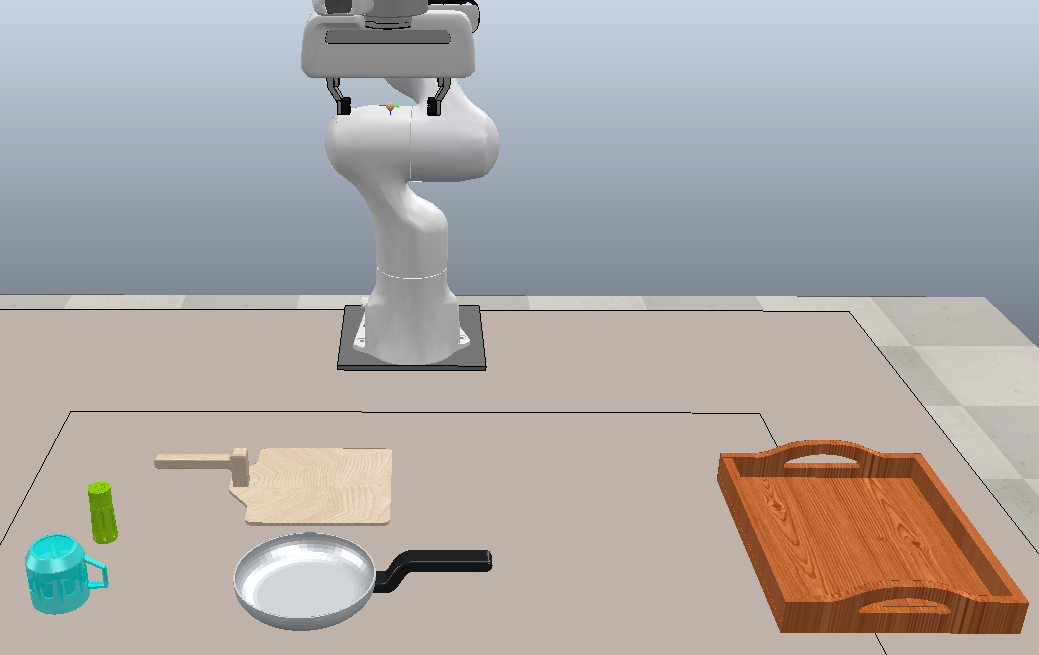}
		\caption{ }
	\end{subfigure}
    \hfill
	\begin{subfigure}[t!]{0.350\columnwidth}
		\includegraphics[width=\columnwidth]{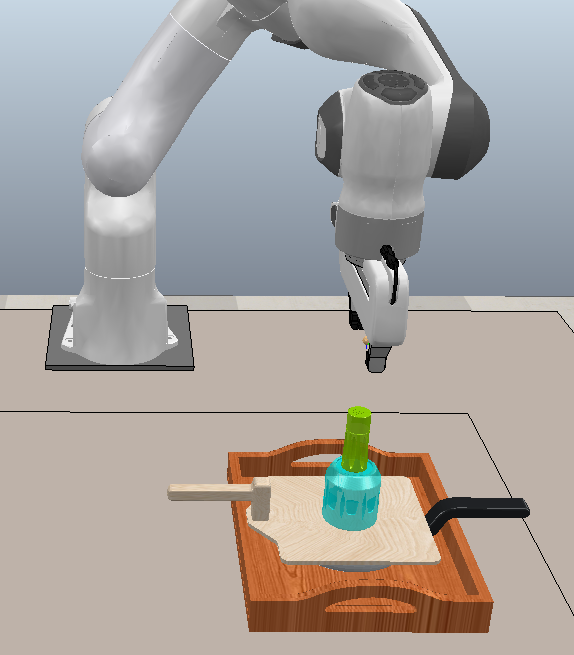}
		\caption{ }
	\end{subfigure}
	\caption{Simulation scenario with various objects used in this study: a salt shaker, a glass mug, a cutting board, a frying pan, and a tray; (a) example initial state; (b) goal state}
	\label{fig:setup}
\end{figure}

In real-world, contact-rich robotic tasks, objects exhibit diverse shapes, sizes, materials, and varying capabilities to interact in specific ways. Such properties can define inter-object affordances, i.e., what can be done to a part of an object with a specific part of another object, as well as physical interaction constraints, e.g., which object can be cut, contained in, or placed on another object.
We propose a TAMP approach that accounts for inter-object affordances for efficient planning and execution of contact-rich tasks in realistic settings. To this end, we use the \emph{Unified} TAMP (U-TAMP) approach~\cite{agostini2023unified} that provides a framework to account for such inter-object affordances in task planning.
U-TAMP defines object-centric abstractions of motion constraints already at the task planning level to ground symbolic actions with minimal sub-symbolic reasoning. However, U-TAMP does not incorporate physical properties or affordances when defining such constraints. We extend these constraints to account for inter-object affordances.

We consider a \textit{kitchenware organization} task as an example of a contact-rich robotic stacking and manipulation task involving real-life objects with various shapes, sizes, and materials (Fig.~\ref{fig:setup}).  
Based on part-based geometric and mechanical properties of the objects, we define abstractions of inter-object affordances and constraints related to grasping, supporting, lifting, and sliding, among others. These abstractions are included in actions such as picking, placement, and sliding in the symbolic realm in such a way that most of the inter-object affordances and the resulting interaction constraints are accounted for.
To assess the efficacy of considering inter-object affordances in long-horizon planning, we additionally solve the planning problem using a Vision Language Model (VLM), which is gaining traction in current TAMP research~\cite{Ding2023, Wake2023, Birr2024, Chen2024, Guo2024a,  Paulius2024b, Wang2024, Zhou2024}.
Furthermore, we also compare our approach with the original U-TAMP~\cite{agostini2023unified}.
In a nutshell, the contributions of this study are as follows:
    1) We account for physical properties of objects (shape, size, material) to define part-based physical interaction constraints in terms of inter-object affordances;
    2) We build robust symbolic abstractions of said affordances related to grasping, supporting, and sliding, to formulate constraints in executing contact-rich pick-and-place and sliding actions;
    3) We define a PDDL-based~\cite{McDermott1998} planning domain that accounts for inter-object affordances and physical constraints to leverage the efficiency of task planners to generate long-horizon feasible plans.

\section{Related Work}
    \label{sec:litrev}

The main focus of traditional TAMP approaches is to define sequences of actions and geometric and kinematics constraints to execute long-horizon tasks ~\cite{Bidot2017, Orthey2024, Vu2024,garrett2018ffrob, garrett2020pddlstream, agostini2023unified}. 
These constraints are usually defined over simplified object models, where objects are represented as cuboids, all of which are light and small enough to be manipulated by the robot and be placed on top of each other, and so forth. These assumptions greatly limit their applicability in real settings. 

Due to recent advances in training deep foundation models on large sets of texts or images, Large Language Models (LLMs) and multimodal Vision-Language Models (VLMs) have found their way into TAMP research. 
These approaches leverage the common sense knowledge and reasoning capabilities of the LLMs to delegate some parts of TAMP to them. 
For instance, Language Models (LMs) have been used for extracting affordances from objects~\cite{Birr2024}, generating symbolic goal descriptions based on the scene information~\cite{Paulius2024b}, performing task planning~\cite{Ding2023, Wake2023, Chen2024, Wang2024, Zhou2024}, determining constraints~\cite{Guo2024a}, or a combination of the above~\cite{Ahn2022}. In particular, SayCan~\cite{Ahn2022} combines an LM with an affordance detector to account for physical constraints for task plan generation.
Use of LMs has resulted in limited performance in realistic settings.
Despite achieving improving success rates in simple manipulation tasks with simple objects, LMs mostly lack the human-level reasoning required to understand all physical interaction constraints and inter-object affordances among objects, and perform reliable long-horizon planning in real environments.

Some other works leverage affordance information in TAMP. Curtis et al.\cite{Curtis2022} use visual perception to estimate affordances that are considered in a feedback loop containing task planning, execution, and monitoring. Ding et al.~\cite{Ding2023} used an LLM to guess valid scene configurations and estimate affordances and constraints to mitigate motion planning cost. Chen et al.~\cite{Chen2024} also leveraged LLMs to estimate affordances or constraints implicitly before planning. In the same line, Cheng et al.~\cite{Cheng2024} defined affordance prompting for LLM to predict action effects and estimate affordances before task planning. These approaches do not build abstractions of affordances for task planning to leverage the efficiency of AI planning to generate physically feasible plans.

Using object-centric representations of constraints and affordances has proven to be useful to articulate task and motion planning; Agostini et al.~\cite{Agostini2015} codified action affordances such as \texttt{cutObj(cucumber), cutWith(knife)} to denote, e.g., that a cucumber can be cut with a knife, so they can be considered at task planning for the execution of long-horizon tasks in complex kitchen settings.
In a later study, Agostini et el.~\cite{Agostini2020} proposed a set of object-centric predicates to represent motion constraints to manipulate individual objects considering motion dependencies between consecutive actions in a plan. 
Inspired by \texttt{StablePick} and \texttt{StablePlace} motion constraint definitions set forth by Garret et al.~\cite{Garrett2021}, Agostini and Piater~\cite{agostini2023unified} proposed a set of grasp and placement constraints in terms of object-centric abstractions that are included in preconditions and effects of picking and placing actions. This allows considering them at task planning level, reducing the need for sub-symbolic reasoning for task execution.
Building on this paradigm, we develop symbolic abstractions of inter-object affordances related to grasping and supporting in contact-rich tasks with heterogeneous objects, in order to construct planning domains that consider physical interaction constraints and generate plans that are feasible in realistic environments.

\section{Background}
    \label{sec:background}
This section introduces the foundational concepts and notation used to describe our contributions.

\subsection{Task and Motion Planning}

In TAMP, task planning uses AI planning heuristic search algorithms to define a sequence of symbolic actions, known as {\it task plan}, that transforms an initial symbolic representation of objects' configuration in the scenario, known as \textit{symbolic initial state} $\mathbf{s}_\mathrm{ini}$, into a goal object configuration, $\mathbf{g}$. 
To this end, it makes use of {\it planning operators} that characterize changes in the symbolic state with actions in terms of 
\textit{preconditions}, which defines the conditions and constraints that need to be met to obtain desired changes with actions, and \textit{effects}, which describes changes in the symbolic state after the action execution, encoded as additions and deletions to the symbolic state. The initial state, goal, and planning operators are composed of \textit{predicates}, which are logical functions that take values \texttt{true} or \texttt{false} if some logical conditions are met.
In this study, we use the Planning Domain Definition Language (PDDL)~\cite{McDermott1998}. 
In PDDL-based symbolic planning, predicates determine object state, properties, and relations. For instance, ``the cup is on the table'' can be encoded as \texttt{\textbf{on} table cup}.
%
After the task plan is generated, a motion planner is used to transform each symbolic action into a feasible trajectory for its execution.
%

\subsection{U-TAMP}
    \label{sec:utamp1}
    
Unified TAMP (U-TAMP)~\cite{agostini2023unified} is a TAMP approach that includes abstractions of motion constraints in the preconditions and effects of planning operators. This permits considering a rich set of motion constraints in the heuristic search of task plans, which, in other cases, would be necessary to carry out during motion planning. As a consequence, the role of the motion planner is effectively minimized, i.e., motion planning is \emph{unified} with task planning.
In the U-TAMP~\cite{agostini2023unified} approach,
objects represent real or virtual objects in the scene, while parts represent the 6 sides of a bounding box, as well as its inside.
%

U-TAMP takes inspiration from Garrett et al.~\cite{Garrett2021} to define a set of constraints for stable grasping, \texttt{\textbf{StableGrasp}}, and stable placement, \texttt{\textbf{StablePlace}}, in picking and placing actions, which are denoted as \texttt{pick-support} and \texttt{place-support}, respectively~\cite{agostini2023unified}.
These constraints are defined as a finite set of legal, scenario-specific, physically plausible interactions between object-object and robot-object parts represented from an object-centered perspective. This formulation enables their use in the heuristic search of task planning and allows symbolic predicates to be mapped directly to object and motion parameters for task plan execution without further deliberation.
For example, these constraints capture legal grasping configurations of an object, its feasible placement orientations on a support, and possible inter-object interactions that can be used in planning. Constraints such as the fact that a surface must be empty before placing an object on it, an object must not be supporting another object on top, so it can be grasped and picked, etc., are symbolically encoded as predicates in the above constraints.
Detailed explanation of the grasping and placement constraints in U-TAMP is given in~\cite{agostini2023unified}. In this work, we enrich the set of \texttt{\textbf{StableGrasp}} and \texttt{\textbf{StablePlace}} constraints to account for object physical properties such as shape, size, and material in terms of inter-object affordances (see Table \ref{tab:constraints}).
Table~\ref{tab:predicates} summarizes the predicates used in this study, along with their description.

\section{U-TAMP using Inter-object Affordances}
    \label{sec:utamp2}

Real-life objects possess physical attributes stemming from their shapes, sizes, or materials, which define their action affordances. For instance, in many circumstances, fragile objects cannot support heavy objects, while large objects cannot be grasped by their body. Many objects can support smaller objects, but not much larger objects stably. Round surfaces are often unsuitable for supporting and placement. 
In summary, inter-object grasping or supporting affordances, which depend on relative properties of objects with respect to each other, create physical interaction constraints that ought to be accounted for in realistic contact-rich manipulation tasks.
Since such properties are not considered in the original U-TAMP, it will lead to unfeasible actions (e.g., grasping a large object, lifting a heavy object) or unstable configurations (e.g., placing a large object on top of a narrow surface or placing an object on a round surface) in realistic settings.

We propose a method that leverages U-TAMP to execute contact-rich robotic tasks involving interactions among objects with heterogeneous shapes, sizes, and materials. This is done by defining grasping, placements, sliding, and lifting constraints that account for inter-object affordances.
In our context, \textit{inter-object affordances} represent the capability of objects and their parts to perform actions with different parts of each other, determined by their physical or functional properties.
Below we list the affordances defined based on our assumptions of physical properties in our particular scenario of kitchen table organization (Fig. \ref{fig:setup}). However, these definitions are generalizable to other tasks and environments as well. 
Note that \textit{object} here (denoted as, e.g., \texttt{?obj1}) refers to any manipulable or virtual object, which could also be a compound object containing many members, \textit{member} (denoted as \texttt{?o1-mem}) refers to an object that is a member of another compound object (e.g., edge, handle, body), and \textit{part} refers to any of the 6 sides of a bounding box along with its inside. In our approach, for a compound object \texttt{?obj1} (e.g., \texttt{frying-pan}), we define the bounding boxes of all of its child members, denoted as \texttt{?o1-mem} (e.g., \texttt{frying-pan-handle}), such that their bounding boxes are parallel to that of the parent object \texttt{?obj1}. 
Therefore, in the context of grasping, we use \texttt{?o1m-hp}, \texttt{?o1m-hf1}, and \texttt{?o1m-hf2} to denote parts of the member object \texttt{?o1-mem} interacting with the gripper.
The object-centric and inter-object affordances introduced in this study are defined as follows.

\textbf{Supportiveness}: Part \texttt{?part1} of object \texttt{?obj1} is \textit{supportive} if it is a flat surface or can behave as one (see \texttt{\textbf{isSupportive}} in Table~\ref{tab:predicates}). 
For example, round surfaces might not be suitable for supporting other objects stably: The \texttt{on} part of a salt shaker is round and not supportive, while the \texttt{on} part of a cutting board is flat and supportive.

\textbf{Supportability}: Similarly, part \texttt{?part1} of object \texttt{?obj1} is \textit{supportable} if it is flat, or can be placed on flat surfaces (See predicate \texttt{\textbf{isSupportable}} in Table~\ref{tab:predicates}). As before, round, thin, or pointy parts of objects might be unsuitable for placement. The two wide parts \texttt{left} and \texttt{right} of a knife, for instance, are supportable, while the pointy \texttt{on}, rounded \texttt{under}, and thin \texttt{front} and \texttt{back} parts are not.

\textbf{Inter-object Supportiveness}: If the shape, size, and material of objects \texttt{?obj1} and \texttt{?obj2} are such that part \texttt{?o2-o1} of \texttt{?obj2} can support \texttt{?obj1} placed on its part \texttt{?o1-o2}, then part \texttt{?o2-o1} of \texttt{?obj2} is \textit{supportive with respect to} part \texttt{?o1-o2} of \texttt{?obj1} (See predicate \texttt{\textbf{isP2PSupportive}} in Table~\ref{tab:predicates}). For instance, a mug can be placed upright on its \texttt{under} side on the \texttt{on} part of a tray. The \texttt{on} part of the tray is therefore supportive with respect to the \texttt{under} part of the mug.

\textbf{Inter-object Graspability}: If object \texttt{?o1-mem}, which is a rigidly-attached member of the compound object \texttt{?obj1}, with its bounding box parallel to that of \texttt{?obj1}, is capable of being grasped stably in such a way that parts \texttt{?o1m-hp}, \texttt{?o1m-hf1}, and \texttt{?o1m-hf2} of the member object interact with the palm, finger1, and finger2 of the gripper, respectively, then \texttt{?obj1} is \textit{graspable} from \texttt{?o1-mem} in this particular grasping orientation (see predicate \texttt{\textbf{canGraspFrom}} in Table~\ref{tab:predicates}). For example, let us consider a frying pan (a compound object) such that its \texttt{on} is the part that contains objects when cooking, \texttt{under} is its bottom placed on the stove, and its handle (a member object) is located on its \texttt{back}. Assuming the handle's bounding box is parallel to the pan itself, the pan is graspable via the handle such that its \texttt{on}, \texttt{left}, and \texttt{right} interact with the palm, finger 1, and finger 2 of the gripper, respectively. The body of the pan, however, is not graspable because it is too bulky to fit within the gripper fingers.

\textbf{Lifting}: If object \texttt{?obj1} is significantly lighter than the payload of the robot, so it can be picked up via its member object \texttt{?o1-mem} in at least one grasping configuration, then \texttt{?obj1} is \textit{liftable} when grasped via \texttt{?o1-mem} (See predicate \texttt{\textbf{canLift}} in Table~\ref{tab:predicates}). For example, a frying pan can be lifted stably when grasped firmly from its handle (handle-grasping). If we try to grasp the pan from the edge of its body (edge-grasping), however, there will likely not be sufficient contact force and friction to tolerate its weight.

\textbf{Sliding}: If object \texttt{?obj1} can be placed on its \texttt{?o1-o2} part and slid stably across the surface \texttt{?o2-o1} of support object \texttt{?obj2}, then \texttt{?obj1} is \textit{slidable} across \texttt{?o2-o1} of \texttt{?obj2} if placed on its \texttt{?o1-o2} (See predicate \texttt{\textbf{canSlide}} in Table~\ref{tab:predicates}). For example, a wooden tray placed on its \texttt{under} part on a kitchen table's \texttt{on} part is slidable across its surface.

\begin{figure}[t]
    \myVspace
	\centering
	\begin{subfigure}[t!]{0.305\columnwidth}
		\includegraphics[width=\columnwidth]{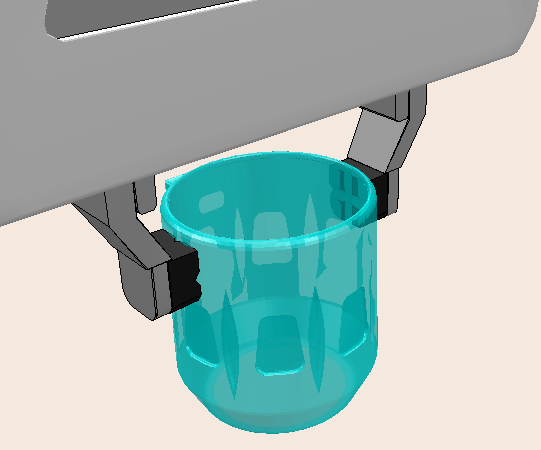}
		\caption{wrap-grasping}
	\end{subfigure}
    \hfill
	\begin{subfigure}[t!]{0.305\columnwidth}
		\includegraphics[width=\columnwidth]{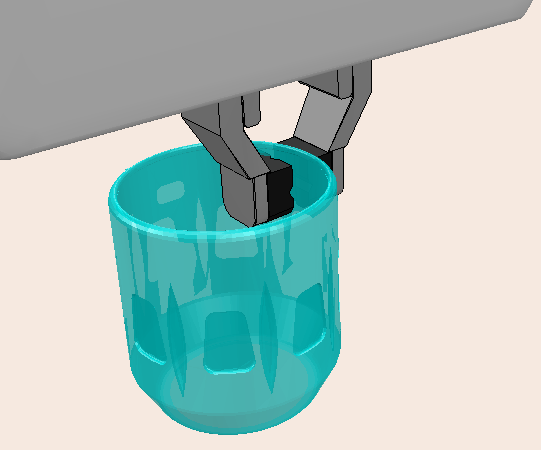}
		\caption{edge-grasping}
	\end{subfigure}
    \hfill
	\begin{subfigure}[t!]{0.35\columnwidth}
		\includegraphics[width=\columnwidth]{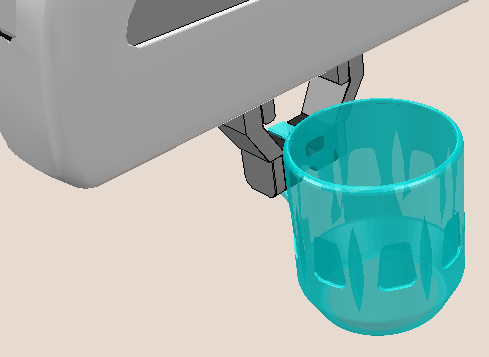}
		\caption{handle-grasping}
	\end{subfigure}
	\caption{The three types of grasping for a vast variety of objects.}
	\label{fig:grasping}
\end{figure}

Predicates representing the above affordances can be seen in Table~\ref{tab:predicates}.
%
We categorize \textbf{grasping} affordances into \textbf{wrap-grasping}, \textbf{edge-grasping}, and \textbf{handle-grasping} (Fig.~\ref{fig:grasping}), distinguished by using different member objects (e.g., \texttt{mug-body}, \texttt{mug-handle}, and \texttt{mug-right-edge}) of the main compound object (\texttt{mug}). In a more general sense, various actions can also require grasping of different members of objects to be executed stably.

\begin{table}[t!]
\centering
\caption{Predicates used in our PDDL domain. Boldface words are names of predicates, and following variables starting with `?' are symbolic arguments. \texttt{?obj} is an object, \texttt{?o1-mem} is its member object, and \texttt{?part} is one of the parts of the bounding box of \texttt{?obj}.}
\label{tab:predicates}
\resizebox{\columnwidth}{!}{%
\begin{tabular}{l|l}
\textbf{Predicate} &
  \textbf{Meaning} \\ \hline
\texttt{\textbf{isObjEqual} ?obj1 ?obj2} &
  \texttt{?obj1} and \texttt{?obj2} are the same object. \\ \hline
\texttt{\textbf{isPartEqual} ?part1 ?part2} &
  \texttt{?part1} and \texttt{?part2} are the same part. \\ \hline
\texttt{\textbf{oc} ?o1-o2 ?obj1 ?obj2} &
  \begin{tabular}[c]{@{}l@{}}Part \texttt{?o1-o2} of \texttt{?obj1} is \\ interacting with \texttt{?obj2}.\end{tabular} \\ \hline
\texttt{\textbf{force} ?obj ?part} &
  \begin{tabular}[c]{@{}l@{}}\texttt{?part} of \texttt{?obj} can be a support, \\ e.g., because it is facing against gravity.\end{tabular} \\ \hline
\texttt{\textbf{base} ?obj ?part} &
  \begin{tabular}[c]{@{}l@{}}\texttt{?part} of \texttt{?obj} is its closest part\\ to the base of the robot.\end{tabular} \\ \hline
\texttt{\textbf{isMember} ?obj1 ?o1-mem} &
  \texttt{?o1-mem} is a member of \texttt{?obj1}. \\ \hline
\texttt{\textbf{isOpposite} ?part1 ?part2} &
  \begin{tabular}[c]{@{}l@{}}\texttt{?part1} and \texttt{?part2} are \\ on the opposite sides of a bounding box.\end{tabular} \\ \hline
\texttt{\textbf{isAdjacent} ?part1 ?part2} &
  \begin{tabular}[c]{@{}l@{}}\texttt{?part1} and \texttt{?part2} are\\ on adjacent sides of a bounding box.\end{tabular} \\ \hline
\texttt{\textbf{isSupportive} ?obj ?part} &
  \begin{tabular}[c]{@{}l@{}}\texttt{?part} of \texttt{?obj} can support\\ other objects placed on top of it.\end{tabular} \\ \hline
\texttt{\textbf{isSupportable} ?obj ?part} &
  \begin{tabular}[c]{@{}l@{}}\texttt{?obj} can be placed on \\ other objects on its \texttt{?part}.\end{tabular} \\ \hline
\begin{tabular}[c]{@{}l@{}}\texttt{\textbf{isP2PSupportive} ?part2 ?obj2}\\ \texttt{\qquad ?part1 ?obj1}\end{tabular} &
  \begin{tabular}[c]{@{}l@{}}\texttt{?part2} of \texttt{?obj2} can stably support\\ \texttt{?part1} of \texttt{?obj1}.\end{tabular} \\ \hline
\begin{tabular}[c]{@{}l@{}}\texttt{\textbf{canGraspFrom} ?obj1 ?o1-mem}\\ \texttt{\qquad ?o1m-hp ?o1m-hf1 ?o1m-hf2}\end{tabular} &
  \begin{tabular}[c]{@{}l@{}}\texttt{?obj1} can be grasped from \texttt{?o1-mem} \\ with this orientation. See Sec.~\ref{sec:utamp2}.\end{tabular} \\ \hline
\begin{tabular}[c]{@{}l@{}}\texttt{\textbf{isGraspFrom} ?obj1 ?o1-mem}\\ \texttt{\qquad ?o1m-hp ?o1m-hf1 ?o1m-hf2}\end{tabular} &
  \begin{tabular}[c]{@{}l@{}}\texttt{?obj1} is being grasped from \texttt{?o1-mem}\\ with this orientation.\end{tabular} \\ \hline
\texttt{\textbf{canLift} ?obj1 ?o1-mem} &
  \begin{tabular}[c]{@{}l@{}}\texttt{?obj1} can be lifted by the robot\\ if grasped from \texttt{?o1-mem}.\end{tabular} \\ \hline
\begin{tabular}[c]{@{}l@{}}\texttt{\textbf{canSlide} ?obj1 ?obj2}\\ \texttt{\qquad ?o1-o2 ?o2-o1}\end{tabular} &
  \begin{tabular}[c]{@{}l@{}}\texttt{?obj1} can be slid on its part \texttt{?o1-o2}\\ on part \texttt{?o2-o1} of \texttt{?obj2}.\end{tabular} \\ \hline
\end{tabular}%
}
\end{table}

\begin{table}[t!]
\myVspace
\centering
\caption{Example \texttt{\textbf{StableGrasp}}, \texttt{\textbf{StablePick}}, \texttt{\textbf{StablePlace}} and \texttt{\textbf{StableSlide}} constraints defined in this study.}
\label{tab:constraints}
\resizebox{\columnwidth}{!}{%
\begin{tabular}{ll}
\texttt{\textbf{StableGrasp[?obj1,?o1-mem,hand]}}        & \texttt{\textbf{StablePlace[?obj1,?obj2]}}                   \\
\texttt{\, (\textbf{isMember} ?obj1 ?o1-mem)}           & \texttt{\, (\textbf{isOpposite} ?o1-o2 ?o1-force)}        \\
\texttt{\, (\textbf{oc} in workspace ?o1-mem)}       & \texttt{\, (\textbf{oc} ?o1-o2 ?obj1 air)}               \\
\texttt{\, (\textbf{base} ?obj1 ?o1-base)}           & \texttt{\, (\textbf{isSupportable} ?obj1 ?o1-o2)}        \\
\texttt{\, (\textbf{oc} ?o1-hp ?obj1 air)}           & \texttt{\, (\textbf{isAdjacent} ?o1-base ?o1-o2)}        \\
\texttt{\, (\textbf{oc} ?o1-hf1 ?obj1 air)}          & \texttt{\, (\textbf{oc} in workspace ?obj2)}             \\
\texttt{\, (\textbf{oc} ?o1-hf2 ?obj1 air)}          & \texttt{\, (\textbf{base} ?obj2 ?o2-base)}               \\
\texttt{\, (\textbf{canGraspFrom} ?obj1 ?o1-mem}        & \texttt{\, (\textbf{isAdjacent} ?o2-base ?o2-o1)}         \\
\texttt{\, \, ?o1-hp ?o1-hf1 ?o1-hf2)}           & \texttt{\, (\textbf{oc} ?o1-o1 ?obj2 air)}               \\
\texttt{\, (not (\textbf{isGraspFrom} ?obj1 ?o1-mem} & \texttt{\, (\textbf{force} ?obj2 ?o2-o1)}                \\
\texttt{\, \, ?o1-hp ?o1-hf1 ?o1-hf2))}          & \texttt{\, (\textbf{isSupportive} ?obj2 ?o2-o1)}         \\
\texttt{\, (not (\textbf{isOpposite} ?o1-base ?o1-hp))} & \texttt{\, (\textbf{isP2PSupportive} ?o2-o1 ?obj2}        \\
\texttt{\, (not (\textbf{isPartEqual} ?o1-hp in))}   & \texttt{\, \, ?o1-o2 ?obj1)}                         \\
\texttt{\, (\textbf{oc} in hand air)}                & \\
                                                     & \texttt{\textbf{StableSlide[?obj1,?o1-mem,?obj2,?space]}}    \\
\texttt{\textbf{StablePick[?obj1,?o1-mem,hand]}}         & \texttt{\, (not(\textbf{canLift} ?obj1 ?o1-mem))}        \\
\texttt{\, (\textbf{StableGrasp}[?obj1,?o1-mem,hand])}  & \texttt{\, (\textbf{canSlide} ?obj1 ?obj2 ?o1-o2 ?o2-o1)} \\
\texttt{\, (\textbf{canLift} ?obj1 ?o1-mem)}            & \texttt{\, (\textbf{isAdjacent} ?o1-space ?o1-o2)}        \\
                                                         & \texttt{\, (not(\textbf{isAdjacent} ?o1-space ?o1-hf1))} \\
                                                         & \texttt{\, (not(\textbf{isAdjacent} ?o1-space ?o1-hf2))} \\
                                                         & \texttt{\, (\textbf{oc} ?o1-space ?obj1 ?space)}         \\
                                                         & \texttt{\, (\textbf{oc} ?o2-o1 ?obj2 ?space)}            \\
                                                         & \texttt{\, (\textbf{oc} in workspace ?space)}            \\
                                                         & \texttt{\, (\textbf{oc} in ?space air)}                  \\
                                                         & \texttt{\, (\textbf{oc} ?space-o2 ?space ?obj2)}         \\
                                                         & \texttt{\, (\textbf{oc} ?space-o1 ?space ?obj1)}         \\
                                                         & \texttt{\, (\textbf{isAdjacent} ?space-o1 ?space-o2)}   
\end{tabular}%
}
\end{table}

\subsection{Actions}

We define three actions: \texttt{pick-support}, for picking an object via the gripper, \texttt{place-support}, for placing an object on top of a surface that can support it, and \texttt{slide-support}, for sliding an object across its support surface. \texttt{pick-support} and \texttt{place-support} are derived from the same actions in the U-TAMP approach \cite{agostini2023unified}. We also introduce the \texttt{slide-support} action as an adaptation of the \texttt{pick-support} and \texttt{place-support} actions for sliding objects on surfaces.
For picking and placing actions, we extend the \texttt{\textbf{StableGrasp}} and \texttt{\textbf{StablePlace}} constraints of the original U-TAMP to account for inter-object affordances and define two new sets of constraints for stably lifting and sliding an object, \texttt{\textbf{StablePick}} and \texttt{\textbf{StableSlide}}, respectively. All these constraints are summarized in Table~\ref{tab:constraints}.
In addition to objects and parts, in our notation \texttt{?space} denotes a virtual space object representing an empty space, \texttt{?obj1-space} denotes the part of \texttt{?obj1} interacting with \texttt{?space}, as well as other similar notations.

In the \texttt{pick-support} action, \texttt{\textbf{StableGrasp}} requires that the object can be stably grasped via its designated member (handle, edge, or body). This is implemented with the predicate \texttt{\textbf{canGraspFrom}}, which accounts for valid grasping configurations of objects' members, and \texttt{\textbf{isGraspFrom}}, which checks the current grasping state (see Table~\ref{tab:predicates}). 
The object's graspable member must be engulfed in the reachable workspace of the robot, checked with the predicate \texttt{\textbf{oc} in workspace ?o1-mem}.
\texttt{\textbf{StablePick}} includes \texttt{\textbf{StableGrasp}} constraints, and additionally requires that the object is light enough to be lifted by the robot when grasped from one of its members.
\texttt{\textbf{StablePlace}} requires that the object is placed stably on its support, but also requires that the support object is fully engulfed in the workspace, and that the interacting parts of the placed object and its support permit a stable placement, checked with the predicate \texttt{\textbf{isP2PSupportive}}.

In the \texttt{place-support} action, \texttt{\textbf{StablePick}} requires that the hand is stably holding the object from the designated three parts of its grasped member, accounted through the predicates \texttt{\textbf{isGraspFrom}} and \texttt{\textbf{canGraspFrom}}. 
\texttt{\textbf{StablePlace}} is as described before, except that it requires that the supportive part of the support and the supportable part of the carried object are clear.  

The \texttt{slide-support} action requires the \texttt{\textbf{StableGrasp}} constraint as described above, but not the \texttt{\textbf{StablePick}} constraint. On the contrary, this action requires that the object \textit{cannot} be lifted by the hand (otherwise \texttt{pick-support} takes priority).  
This action includes the \texttt{\textbf{StableSlide}} constraints, which require that the object can be slid on its current support surface, checked by \texttt{\textbf{canSlide}} predicate, and that there is an empty space adjacent to the object's bounding box lying on the object's support surface, and can fully contain the object.
As a result of this action, the hand is free, the object is still lying on the support, but has been moved inside the space, which now contains the object.
The complete PDDL problem and domain definition files are available at \url{https://osf.io/egrjv/overview?view_only=05ceb38e4ce2432699a53975bcafa022}.

\section{Experiments}
    \label{sec:exp}

Our aim is to achieve higher planning success rates while maintaining a fast planning time. We facilitate this by enriching the planning domain using abstractions of affordances, thereby minimizing sub-symbolic reasoning.
For the experimental assessment, we use a kitchen organization task (Fig.~\ref{fig:setup}), where the goal is to stack different subsets of objects on a tray. In our scenario, the following objects are used: \textbf{Salt shaker:} This lightweight, cylindrical object cannot support other objects stably on its round head, and can only be placed stably on its bottom side. It can only be grasped stably from its top side. \textbf{Glass mug:} It can support small objects, like the salt shaker, but not large objects on top of it, due to instability in placement. It cannot be placed on any of its round lateral sides, because it will roll off, but can be placed on its top and bottom sides stably. It can be wrap-grasped from its top and bottom sides, and it can also be handle-grasped. \textbf{Cutting board:} This is a flat surface equipped with a handle for the robot to grasp. It is structurally robust and capable of supporting other objects on its top flat surface, but it cannot be placed upside-down. It can only be handle-grasped. \textbf{Frying pan:} It can only be placed on its bottom side stably, and can only be handle-grasped. It is too large to be wrap-grasped and too heavy to be edge-grasped from any of its body's edges. It can only support the cutting board on its top side. \textbf{Wooden tray:} The tray is a large and heavy object that cannot be lifted off the table, but can be slid across its surface. It can be grasped from handles and edges. It is large and robust enough to support all other objects on its top, and can only be placed upright.

The scenario is implemented in the physically realistic simulator CoppeliaSim~\cite{coppeliaSim}. We define a scene including a Franka Emika Panda robot with the two-finger Panda gripper installed on a kitchen table, along with 5 objects used in our experimental configurations.
In the initial state, the objects are scattered around the table in various configurations where some objects are stacked on others, and the goal is to \emph{organize} the kitchen table by stacking all existing objects on top of one another on the tray.
Containment and supporting are strictly distinguished in our scenario. For example, even though a bowl can \emph{contain} an ice cube, it cannot \emph{support} an ice cube on top; the cube will fall inside. This definition affects the supporting affordances and planning. 
In order to assess the capabilities of the approaches, we create a list of 127 valid combinations of stacks of objects in the scene, while sweeping the list of manipulable objects from 2 to 5. 
We evaluate planning computation time and success rate for different scene complexities.

\subsection{Task Planners and Evaluation Metrics}
\begin{figure}[t!]
    \myVspace
	\includegraphics[width=\columnwidth]{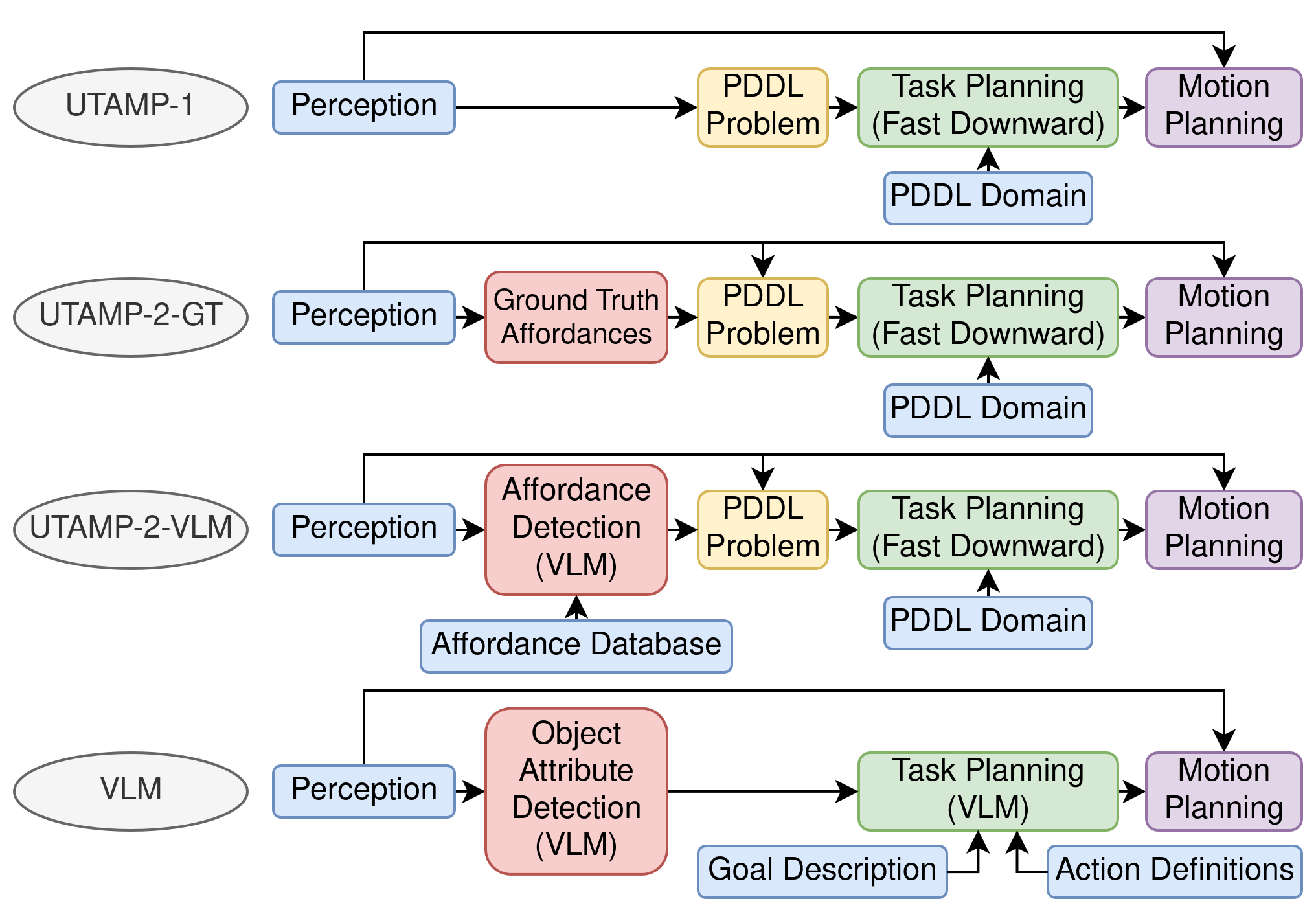}
	\centering
	\caption{Experimental pipeline in the main approaches implemented in this study.}
	\label{fig:pipeline}
\end{figure}

One of the most important responsibilities of the task planner in this study is to account for inter-object affordances and physical interaction constraints to come up with a feasible task plan. We tackle task planning in two different ways, using PDDL-based and VLM-based task planners.   
Fig.~\ref{fig:pipeline} highlights the similarities and differences among the approaches used in our experiments.

\subsubsection{PDDL-based Task Planning} 
In this setting, we extend the planning domain of U-TAMP ~\cite{agostini2023unified} to account for the object-centric inter-object affordances and constraints of Table \ref{tab:constraints}.
To define the PDDL description of the goal, we require that all the slots on the table that hold objects must be clear, except for the one of the tray. We utilize the following approaches based on PDDL. In all of them, we use the off-the-shelf Fast Downward planner \cite{Helmert2006}.

\textbf{I. UTAMP-1:} This is the original Unified TAMP by Agostini et al.~\cite{agostini2023unified}, where object-centric predicates are used to account for motion and kinematic constraints. It does not, however, account for inter-object affordances and physical attributes of objects (Fig.~\ref{fig:pipeline}). 
%
Our proposed method, referred to as \textit{UTAMP-2}, extends the planning domain of UTAMP-1 to account for inter-object affordances (supporting, grasping, lifting, sliding), workspace engulfment preconditions, and the \texttt{slide-support} action for heavy objects (Fig.~\ref{fig:pipeline}). We deploy this method in two approaches.

\textbf{II. UTAMP-2-GT:} This approach is used as a ground truth (GT) baseline, where inter-object affordances are manually hand-crafted to showcase the performance of our proposed method, i.e., using symbolic abstractions of inter-object affordances at task planning level, with perfect affordance detection.

\textbf{III. UTAMP-2-VLM}: In this approach, we utilize a VLM to get a scene image, detect the objects therein, and their physical attributes (shape, size, material, etc.). A parser then transforms these attributes into a list of inter-object affordances and composes them into the PDDL problem.
In the VLM pipeline, we prompt examples of a variable set of objects and attributes. Then, we provide the scene image captured by a vision sensor in the CoppeliaSim scene and ask the VLM to detect objects and provide a list of their attributes in a specific format.
    
\subsubsection{VLM-based Task Planning} 
We utilize VLMs to leverage their common-sense knowledge and reasoning skills, implicitly accounting for inter-object affordances to generate physically valid plans. This is done by implementing a proxy of the SayCan~\cite{Ahn2022} approach, in the same line as Paulius et al.~\cite{Paulius2024b}, where the AI assistant is provided with a series of prompts in which we explain its role, provide a scene image, and use it for task planning. When describing the actions to the VLM, the \texttt{pick-support} action is simplified as \texttt{pick ?obj1 ?obj2}, \texttt{place-support} as \texttt{place ?obj1 ?obj2}, and \texttt{slide-support} as \texttt{slide ?obj1 ?o1-space}, to ease the VLM's work in reasoning and task plan composition. 

\textbf{IV. VLM:} First, we prompt the AI assistant's role and a description of all possible actions it can use in planning, along with their preconditions and effects. We also explain the object attributes (shape, size, material, etc.) of example objects relevant for affordance detection. An image of the scene in the initial configuration is provided afterward, and the assistant is asked to detect all objects therein. Then, we ask the assistant to provide all relevant object attributes for all detected objects. This leverages the chain-of-thought reasoning capability of the VLM to improve planning and reasoning later on. Afterward, we explain to the VLM that the goal is to stack all objects on a single object (the tray in our scenario), and ask to generate a PDDL-compatible task plan (Fig.~\ref{fig:pipeline}). To generate a valid plan, the VLM needs to implicitly reason about inter-object affordances and physical interaction constraints. 
We use the \texttt{o4-mini} model of OpenAI's ChatGPT, which has advanced reasoning capabilities.
A complete prompting example is available at \url{https://osf.io/egrjv/overview?view_only=05ceb38e4ce2432699a53975bcafa022}.
%

For all the approaches above, after generating the task plan, relevant symbols are decoded to real values, where Cartesian end-effector poses for grasping, placement, or sliding are calculated. We adopt a similar approach to the original U-TAMP to derive hand poses from relative object poses and bounding boxes~\cite{agostini2023unified}. For motion planning, we use the RRT-Connect planner~\cite{RRTConnect} available in the \texttt{SimOMPL} plugin in CoppeliaSim.
We evaluate our task planners in terms of \textit{success rate} and \textit{planning time}. 

\textbf{Success rate:} We check for the validity of the task plan using the physical constraints described in Section~\ref{sec:utamp2}. If the necessary physical conditions are met before every action and if the accumulated effects are also compatible with them, the plan is considered as valid. If any of these conditions are not met, the plan is marked as a failure. Otherwise, the plan is executed in a simulated scenario. If all the actions in the symbolically valid task plan are executed without issue, then the task plan has been successful. Otherwise, again, the plan is a failure. Success rate is calculated as the portion of initial configurations that resulted in a successful execution of the task.

\textbf{Planning time:} Task planning computation time is calculated from the moment the planning process starts, up to the moment the task plan is received. Statistics of planning time across a set of initial configurations are reported as mean and standard deviation.

\begin{figure*}[t!]
    \myVspace
	\includegraphics[width=\textwidth]{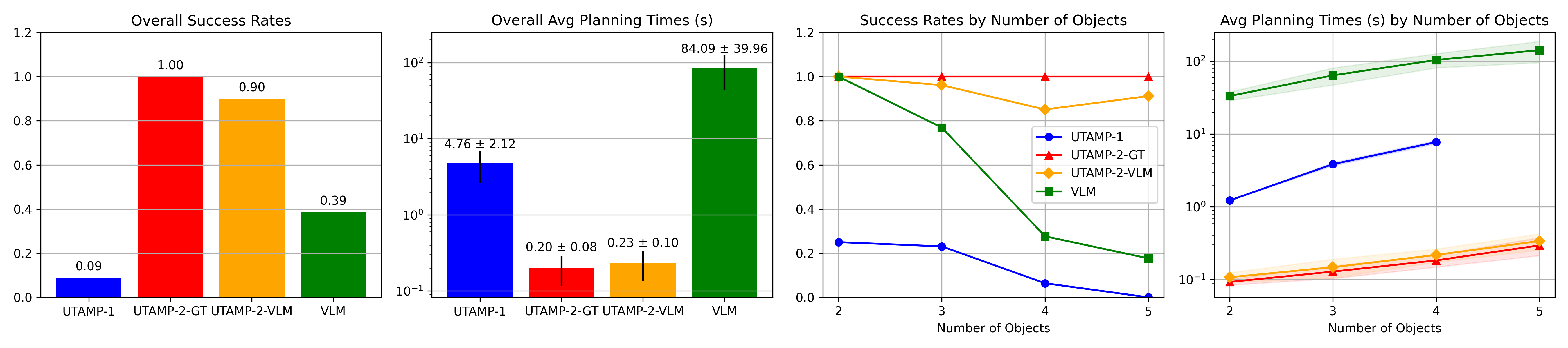}
	\centering
	\caption{Results of experiments only involving pick and place actions. Results show success rates and planning time, the latter in a logarithmic scale, where bold lines are means and shades are standard deviations. The left pair shows the overall results, while the right pair shows the per-object-count results.}
	\label{fig:resultsAll}
\end{figure*}

For the experimental evaluation, we define two different tasks. The first task only considers cases where all objects are engulfed by the reachable workspace, meaning no sliding action is required. This is done to allow assessing performance across all four baselines, including the original UTAMP-1, which does not consider the sliding action in its domain definition. The other task only comprises cases where the wooden tray is not engulfed in the reachable workspace and requires a sliding action first. This task serves for the comparison of UTAMP-2 and VLM approaches in a more complex setting.
We select scenes with $n$ manipulable objects, where $2 \leq n \leq 5$, and sweep them to define initial valid configurations.
We report the statistics of the metrics for each value of $n$ and each planning method, and inspect the results to see how performances change with increasing scene complexity. 
We also show overall success rate and computation time of generated plans for all selected initial configurations for each planning approach. 
The VLM used in our experiments is the \texttt{o4-mini} model of OpenAI's ChatGPT, with a high reasoning effort. 
All experiments were performed on a system with a 6-core 13th-generation Intel i7-1355U CPU, 32 GiB of DDR5 RAM, in Ubuntu 22.04.5 LTS x86-64 OS.

\subsection{Results}
    \label{sec:res}

The left pair of figures in Fig.~\ref{fig:resultsAll} shows the overall performance. It highlights the advantage of using rich symbolic abstractions in task planning, achieving higher success rates and faster planning times. It also shows that when realistic objects with nuanced properties are involved, language models are likely to fail to compete with symbolic planning methods.

In terms of success rates, UTAMP-2-GT, which accounts for ground-truth inter-object affordances and constraints at planning time, maintains the highest success rate of 100\%, while UTAMP-2-VLM, which uses a VLM for affordance detection, achieves the second place at 90\% success rate. The other approaches fall below 50\%. 
To further assess affordance detection performance in the UTAMP-2-VLM approach, we also calculated precision, recall, and F\textsubscript{1} score using ground truth affordances as reference. All three metrics have a mean value of 0.98 in our experiments. Nonetheless, the success rate dropped to 90\% in UTAMP-2-VLM. This is because our task is sensitive to affordance detection performance, and just a few incorrectly assigned supporting affordances make stacking the objects unfeasible, leading to a failure.
As expected, the performance of UTAMP-1 is poorer because, even though it considers object-centric abstractions of basic motion constraints, it does not account for constraints coming from objects' physical properties, e.g., which object can(not) stably support others, which object can(not) be lifted by the robot, etc. 
VLM has a subpar success rate of 39\%, since we are dealing with a complex scenario with realistic objects and nuanced affordances that depend on physical details of parts of objects.
Regarding planning times, UTAMP-2 variants generate plans significantly faster than the other methods, including UTAMP-1, by 2 orders of magnitude on a logarithmic scale. This indicates that the introduced inter-object physical constraints effectively bootstrap the heuristic search in task planning. Action preconditions in UTAMP-1 are not as rich and expressive as UTAMP-2, leading to longer planning times.
VLM-based planning is, on average, two orders of magnitude slower than UTAMP-2 due to the latency and response time of the VLM, especially with high reasoning effort. It also has the highest stochasticity, resulting in the largest variance.

The right pair of figures in Fig.~\ref{fig:resultsAll} shows the \textit{per-object-count} performances. Accordingly, while UTAMP-2 variants maintain the highest success rate, other methods become less successful with increasing scene complexity. Especially, VLM and UTAMP-1, which are the most naive about inter-object affordances and physical constraints, can seldom come up with a successful task plan or a valid stacking order with 5 objects. Planning time increases as expected with increasing scene complexity but UTAMP-2 variants have a tamed slope compared to the others, and do not increase by a full order of magnitude. 
UTAMP-1, however, takes significantly longer for large numbers of objects, further suggesting that the introduced inter-object physical constraints help guide the heuristic search in AI planning more efficiently.
%
\begin{table}[t]
\centering
\caption{Performance of the approaches in configurations that require the sliding action. Planning time is in seconds.}
\label{tab:results-sliding}
\resizebox{\columnwidth}{!}{%
\begin{tabular}{llll}
\textbf{Method} & \textbf{UTAMP-2-GT} & \textbf{UTAMP-2-VLM} & \textbf{VLM} \\\hline
Success rate    & 1.0                 & 0.81                 & 0.41         \\
Planning time   & 0.17 ± 0.05         & 0.20 ± 0.08          & 66.9 ± 25.2 
\end{tabular}%
}
\end{table}

After performing the experiments on all configurations that only require pick and place, we also test the approaches on the initial configurations in which the wooden tray is partially outside the reachable workspace of the robot, and since it is not liftable but is slidable, will require the sliding action to be slid into the workspace of the robot, before stacking other objects on it.
We test all approaches except UTAMP-1 here, because it does not support the sliding action. Table~\ref{tab:results-sliding} shows the results in this experiment. Comparative results hold true as they did before, with VLM having the lowest success rate and the longest planning time, and UTAMP-2-VLM having a lower success rate than U-TAMP-2-GT.

It is important to note that, after close inspection, we verify that all the plans generated by UTAMP-2-VLM are symbolically valid and do not violate any constraints. Failures occur due to the planner finding no feasible plans, mostly due to missing supporting affordances required for stacking, which were not generated due to an inconsistent detection of object attributes by the VLM.
As for the VLM, we observed that planning failures arise mostly due to violations of the support constraints, e.g., placing small objects inside the frying pan or placing an object on an occupied support surface, even though this was explicitly mentioned as invalid cases in the prompting. We also observed violations of picking constraints, where plans comprise two consecutive picking actions, and violations of inter-object affordances to define the correct order for stacking the objects.

\section{Discussions}
PDDL planners use AI symbolic planning to generate plans that satisfy constraints specified in action preconditions. Therefore, not only are they efficient, but the plan they generate is guaranteed to be (symbolically) valid.
Deep learning architectures and LLMs/VLMs inherently lack this feature, making them likely inferior in efficient long-horizon planning. However, they are well-suited for affordance detection or symbol grounding in perception, since the common-sense reasoning skills of LLMs can transcend the limitations of supervised learning or methods that learn from limited interactions or demonstrations.

Our proposed UTAMP-2 planning domain has some limitations despite its performance. Firstly, there are simplifying assumptions and rules in affordance assignment and motion planning. We assume that small objects cannot support larger ones, that round surfaces are unsuitable for supporting/placement, that large objects are always too heavy to be lifted, and so forth. We also assume that the ideal supporting/placement position of a flat surface is invariably its centroid, and that the ideal grasping position of a member is from the middle. These assumptions are valid for a wide variety of objects and scenarios, but not all of them. 
In the future, we aim to address these limitations and extend robust symbolic abstractions to more complex actions.

\section{Conclusions}
We propose a novel PDDL-based planning domain that utilizes abstractions of inter-object affordances to account for physical constraints in grasping, supporting, lifting, and sliding, to provide expressive task plans to execute contact-rich tasks involving realistic objects. We generate symbolic actions in such a way that said constraints are taken into account at task planning (which is fast, efficient, and guarantees constraint satisfaction) and no longer need to be reasoned about during motion planning. 
We leverage the common-sense knowledge of vision-language models (VLMs) to assist in affordance detection, though alternative learning-based approaches can also be employed. Additionally, we provide a hand-crafted set of affordances as a ground-truth reference.
Empirical comparison of our method with the original U-TAMP and VLM-based planners demonstrates that, given robust symbolic abstractions of inter-object affordances and constraints, our approach exploits the efficiency of AI planning to quickly generate feasible and expressive task plans without the need for sub-symbolic reasoning.

\bibliography{references}




\end{document}